\documentclass[letterpaper]{article} 
\usepackage{aaai24}  
\usepackage{times}  
\usepackage{helvet}  
\usepackage{courier}  
\usepackage[hyphens]{url}  
\usepackage{graphicx} 

\usepackage{latexsym}
\usepackage{amssymb}
\usepackage{amsmath}
\usepackage{amsthm}
\usepackage{booktabs}
\usepackage{enumitem}
\usepackage{color}

\usepackage{array}

\usepackage{eqparbox}
\usepackage{url}
\usepackage{multirow}
\usepackage{tabularx}

\usepackage{booktabs}
\usepackage{diagbox}
\newcolumntype{C}{>{\centering\arraybackslash}X}
\usepackage{adjustbox}
\usepackage{pifont}
\usepackage{mdwmath}
\usepackage{mdwtab}
\usepackage{xcolor}
\usepackage{changepage,threeparttable}

\usepackage{natbib}  
\usepackage{caption} 
\usepackage{algorithm}
\usepackage{algorithmic}

\usepackage{newfloat}
\usepackage{listings}
\DeclareCaptionStyle{ruled}{labelfont=normalfont,labelsep=colon,strut=off} 
\floatstyle{ruled}
\newfloat{listing}{tb}{lst}{}
\floatname{listing}{Listing}
\title{
S-RAF: A Simulation-Based Robustness Assessment Framework for Responsible Autonomous Driving}
\author{
   Daniel Omeiza\textsuperscript{\rm 1}\thanks{Corresponding author.},
    Pratik Somaiya\textsuperscript{\rm 1},
    Jo-Ann Pattinson\textsuperscript{\rm 3},
    Carolyn Ten-Holter\textsuperscript{\rm 1},
    Jack Stilgoe,
    Marina Jirotka\textsuperscript{\rm 1},
    Lars Kunze\textsuperscript{\rm 1,2}
}

\affiliations {
    \textsuperscript{\rm 1}University of Oxford, UK,\\
    \textsuperscript{\rm 2}University of the West of England,\\
    \textsuperscript{\rm 3}University of Leeds,\\
    \textsuperscript{\rm 4}University College London\\
    \{danielomeiza, pratik, lars\}@robots.ox.ac.uk, 
    carolyn@cs.ox.ac.uk,
    j.m.pattinson@leeds.ac.uk,
    marina.jirotka@cs.ox.ac.uk,
    j.stilgoe@ucl.ac.uk
}

\usepackage{bibentry}

\begin{document}

\maketitle
\begin{abstract}
As artificial intelligence (AI) technology advances, ensuring the robustness and safety of AI-driven systems has become paramount. However, varying perceptions of robustness among AI developers create misaligned evaluation metrics, complicating the assessment and certification of safety-critical and complex AI systems such as autonomous driving (AD) agents. To address this challenge, we introduce Simulation-Based Robustness Assessment Framework (S-RAF) for autonomous driving. S-RAF leverages the CARLA Driving simulator to rigorously assess AD agents across diverse conditions, including faulty sensors, environmental changes, and complex traffic situations. By quantifying robustness and its relationship with other safety-critical factors, such as carbon emissions, S-RAF aids developers and stakeholders in building safe and responsible driving agents, and streamlining safety certification processes. Furthermore, S-RAF offers significant advantages, such as reduced testing costs, and the ability to explore edge cases that may be unsafe to test in the real world. Code for this framework is available\footnote{https://github.com/cognitive-robots/rai-leaderboard}
\end{abstract}

\section {Introduction}
Responsible AI (RAI) development has gained increased attention lately as the development and application of sophisticated AI technologies continue to increase immeasurably with threats of harm to society~\cite{allen2024real}. Beyond the assessment of technical capabilities (e.g., accuracy), the potential threats these technologies pose highlight the need for a more collective assessment of their \textit{purpose} and \textit{process} to mitigate associated risks while harnessing their potential for good. Of late, deep tech corporations, e.g., autonomous vehicle companies, seem to operate in silos regarding the communication of new knowledge, safety test cases, and reports. Many develop their own metrics and benchmarks for assessing their technologies, thus, making it difficult for regulators and other parties of interest to have a fair assessment of these technologies across board. We, therefore, advocate for the development of accessible frameworks for easier and more objective assessment of safety-critical responsible AI principles such as robustness, and environmental sustainability, among others.

We conceptualise RAI as the conscious effort in designing, developing, and deploying artificial intelligence (AI) systems in a way that maximises their benefits while minimising their risks to people and society at large.

\begin{figure}[h!]
\centering
  \includegraphics[width=\columnwidth]{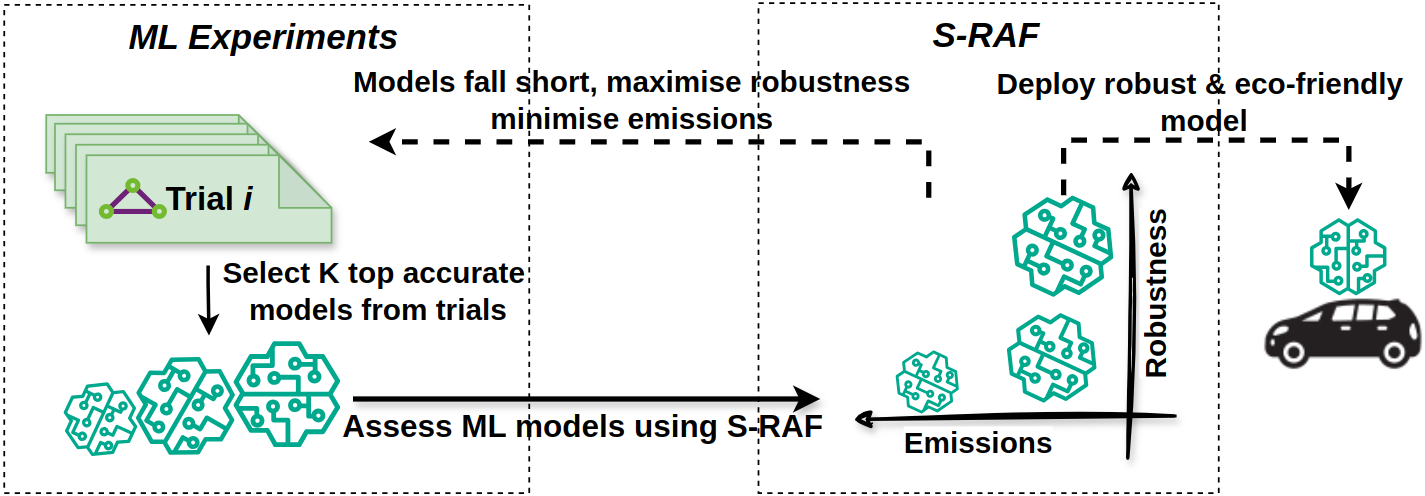}
  \caption{ \small{Overview of S-RAF. Trained agents/models from the ML trials are selected and passed to S-RAF for comprehensive robustness and $\text{CO}_2$ emission assessment, and are ranked accordingly.}}~\label{fig:teaser}
\end{figure}
If we are developing artefacts to act with some autonomy, then `we had better be quite sure that the purpose put into the machine is the purpose which we really desire' as quoted in~\cite{dignum2019responsible}. In this light, the development of a framework for assessing each dimension of RAI emerges as a pivotal tool for \textit{responsible innovation,} a broader subject guiding the development of technologies that align with societal needs and values~\cite{stilgoe2020developing, jirotka2017responsible}. While purposes are very hard to technically account for, processes might be amenable. Thus, we developed a robustness assessment framework (S-RAF) that focuses on assessing how AD agents handle unforeseen situations during operation~\cite{marcus2020next}. 
AD is a safety critical domain with reports of accident cases~\cite{cruise, ekim7}. Hence, the need for responsible innovation, involving rigorous assessment of safety-critical RAI principles such as robustness.

While there might be a finite list of RAI principles in the academic literature, as a first step, we have focused on robustness (a safety critical concept in AD) and examined how efforts to improve robustness impact carbon emission, an indicator for environmental sustainability in this paper.
We tested three state-of-the-art AD models to demonstrate how RAF can be applied in practical settings. This encompasses the assessment of robustness and the investigation of trade-offs with sustainability and transparency (see Fig.~\ref{fig:teaser}).

While there are a few accessible developments around RAI, most of the efforts are limited to guidelines, conceptual frameworks, and toolkits. The existing toolkits are mainly targeted toward regular machine learning models and are impracticable for assessing complex goal-based systems such as AD agents. Moreover, they also mainly support single modality input (either RGB sets or text/audio sequence) and/or require structured input types.
In line with the notion of the `veil of ignorance'~\cite{rawls2020theory}, where impartial decisions are made without knowledge of one’s social position, S-RAF aims to embody a similar principle by ensuring unbiased robustness assessments of complex AD agents.

\section{The Need for RAI Indicators}
We focus on robustness and sustainability indicators in this section.
\paragraph{Robustness} One of the core principles of RAI is robustness~\cite{floridi2021establishing}. In contrast to symbolic AI systems, deep learning models often deployed in AD undergo training with extensive datasets, and their complex structures, which may consist of millions or billions of parameters, are not collectively readily interpretable by humans. Consequently, it is currently impractical to offer comprehensive assurances regarding the accurate performance of neural networks when they encounter input data significantly divergent from what was seen during training. Nonetheless, numerous AI applications (including AD) have critical security or safety implications, necessitating the ability to assess the systems' resilience when confronted with unforeseen events, whether they arise naturally or are deliberately induced by malicious actors~\cite{berghoff2021robustness}. There are many methods of adversarial attacks such as those created using generative models like GAN, VAE~\cite{bowles2018gan}, and those created by adversarial perturbations like LBFGS~\cite{szegedy2013intriguing}, FGSM~\cite{goodfellow2014explaining}, Deep Fool~\cite{moosavi2016deepfool}. On the other hand, multiple works exist addressing this question of robustness against such attacks~\cite{zhang2020interpreting, berghoff2021robustness}. Some works~\cite{huang2017safety, gehr2018ai2, singh2019abstract} have leveraged formal verification to provide robustness guarantees. These methods are faced with the limitations of scalability to the large neural networks often used in practice. Moreover, the defence provided is usually against artificial attacks at the pixel level. Hence, our notion of robustness in this paper is the ability of the AD agent to maintain a consistent behaviour or gracefully handle unforeseen events that might \textbf{naturally} occur in its operating environment. We seek to address questions such as, what happens when a sensor of an autonomous vehicle (AV) malfunctions? What happens when it is occluded by dirt or other materials? What happens when there is a sudden serious decline in weather conditions? These conditions are serious and can lead to fatal accidents when not effectively handled. For instance, when a camera fails, one expects the other cameras to make up for this failure, and do so without constituting safety risk or violating traffic rules. While the primary focus of this paper is on robustness, researchers have drawn connections between robustness and other indicators, especially sustainability. With many AI companies following the scaling law---which suggests that increasing the number of model parameters leads to improved performance---the demand for computational resources continues to grow. This escalating resource consumption consequently results in higher carbon dioxide $\text{CO}_2$ emissions.

\subsubsection{Carbon Emission}
According to Strubell et al.~\cite{strubell2019energy}, the training process of a single deep learning natural language processing (NLP) model on a GPU can result in the emission of approximately $\sim{272},155kg$ of $\text{CO}_2$, comparable to the lifetime emissions of five cars. Similarly, Google's AlphaGo Zero generated 96 metric tonnes of $\text{CO}_2$ during 40 days of training, equivalent to 1,000 hours of air travel or the carbon footprint of 23 American households~\cite{van2021sustainable}. The environmental impact of recent Generative AI models (e.g., GPT-3, ChatGPT, DALL-E, etc.) is even more concerning. Efforts to build more robust, multi-tasking models are observed to negatively impact environmental sustainability. Other than train time emissions, applications like ChatGPT, handling 11 million requests per hour, emit 12.8 thousand metric tons of $\text{CO}_2$ annually, 25 times higher than GPT-3's training emissions~\cite{chien2023reducing}. This has implications for AD, where such models are increasingly integrated.
Research has assessed AI's energy use and $\text{CO}_2$ emissions~\cite{strubell2019energy, lacoste2019quantifying, dhar2020carbon}, focusing on the carbon impact of various Graphics Processing Units (GPUs). We adopt this approach and report $\text{CO}_2$ emissions of benchmark AD agents by tracking their software processes.

\section{Previous RAI Efforts}
\paragraph{RAI Frameworks} The development of RAI framework (including tools) is essential for all the different stages of AI development and deployment. Several efforts have been channelled toward developing RAI guidelines and frameworks over the years. Based on the survey by Berman~et al.~\cite{berman2024scoping}, the efforts include the provision of guidance for problem formulation and procurement decisions for the appropriate AI system for the given use case as equally argued in~\cite{coston2023validity}. It also includes ethical considerations for designing the systems, e.g., checklists~\cite{lifshitz2020legal}, and procedures for enabling participatory design~\cite{gerdes2022participatory}. In the actual machine learning workflow, dataset collection and training processes benefit from established ethical guidelines~\cite{rhem2023ethical}. Some of the frameworks are also useful for model post-training activities, e.g., for fairness assessment~\cite{agarwal2018reductions}, model-card~\cite{mitchell2019model}, and datasheet~\cite{gebru2021datasheets} for model training documentation, dataset details, and model reporting. Another use case is the facilitation of an effective AI system auditing process~\cite{krafft2021action}. These efforts are limited to data-driven models. Other than the CARLA Leaderboard~\cite{leaderboard} which is mainly performance-focused, there seems to be a dearth of accessible frameworks and tools in AD that are focused on RAI principles.
Our work contributes to the efforts by defining metrics for and providing a software tool for assessing the robustness of AD agents and understanding connections with carbon emissions.

\subsubsection{RAI Tools}
RAI tools have been developed for use by AI practitioners, of which a few focus on robustness (e.g.,  ART~\cite{nicolae2018adversarial}, FoolBox~\cite{rauber2017foolbox}, RobustBench~\cite{croce2020robustbench}), explainability (e.g., Google What-if tool~\cite{wexler2019if}, Captum~\cite{kokhlikyan2020captum}, IBM AI Explainability 360~\cite{mojsilovic2019introducing}, etc), a handful on sustainability (e.g., CodeCarbon~\cite{schmidt2021codecarbon} and Eco2AI~\cite{ budennyy2022eco2ai}), fairness (e.g., IBM Fairness 360~\cite{bellamy2019ai}, Fairlearn~\cite{fairlens}, etc). Our work spans robustness, to include  $\text{CO}_2$ emission tracking and supports complex multi-modal AD agents.

In summary, building responsible AD agents requires additional efforts beyond adversarial robustness for models with single input modality, and should be done without trading off environmental sustainability. This is the gap that S-RAF potentially seeks to fill.

\section{S-RAF: Robustness Indicators}

We consider an agent to be robust if it can sustain performance in the presence of environmental disturbances, measurement noise, and data drift without infractions of traffic rules or collisions.

\subsection{Robustness against Environmental Disturbances}
Robustness against environmental disturbances is paramount to ensure the safe and reliable operation of vehicles. There are different types of disturbances peculiar to different sensor types:
\paragraph{i. Camera Occlusion}
Environmental materials such as dirt, leaves, and snow accumulation on sensors, among others can cause camera occlusion. Camera occlusion occurs when the camera's field of view is obstructed, leading to incomplete or inaccurate perception data. In this disturbance situation, we assume that the disturbance occurs directly on the camera lens.

Formally, let \(I_\text{original} (x, y)\) represent the original image captured by the camera, where \(x\) and \(y\) denote the spatial coordinates of the image. Image occlusion can be represented by introducing an occlusion mask \(M (x, y)\) that indicates the regions of the image affected by occlusion. The occluded image \(I_{\text{occluded}}(x, y) \) is defined as:
\[I_{\text{occluded}}(x, y) = I_{\text{original}}(x, y) \cdot (1 - M(x, y))\]

\paragraph{ii. LiDAR Occlusion}
Similar to cameras, occlusion can arise in 3-dimensional (3D) Light Detection and Ranging (LiDAR) from environmental factors, including leaves, snow, dirt, etc. Such environmental elements have the potential to obscure the sensor's enclosure, leading to instances of occlusion where specific regions of the observed scene become inaccessible or exhibit data incompleteness. 

\begin{figure}[h]
\centering
  \includegraphics[width=\columnwidth]{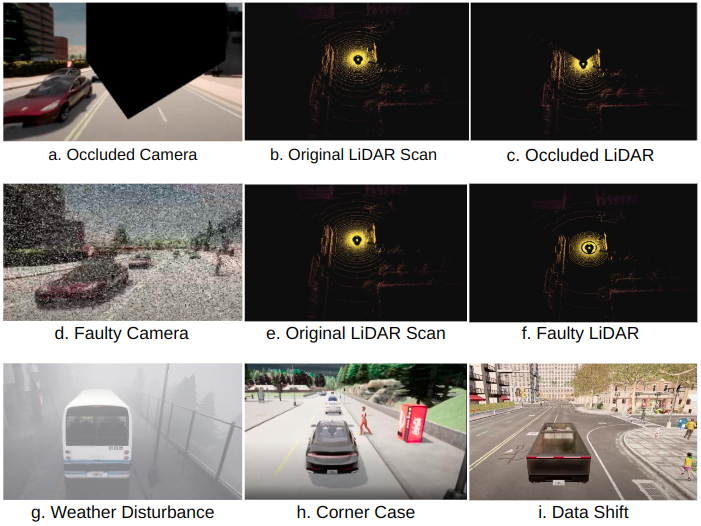}
  
  \caption{ \small The first row shows samples of camera occlusion (Fig. 2a) and LiDAR occlusion (normal: Fig. 2b, occluded: Fig. 2c). 2nd row shows faulty sensors for a camera (Fig. 2d) and for LiDAR (normal: Fig. 2e, faulty with missing rays: Fig. 2f). Third row shows a rainy weather (2g), a jaywalker (Fig. 2h), and data drift example, a chaotic scene (Fig. 2i).}~\label{fig:rai_example}
\end{figure}

If the sensor data is represented as $S(x,y,z)$, where $x$, $y$, and $z$ are spatial coordinates of each LIDAR point, then occluded data can be written as:
\[S_{\text{occluded}}(x, y, z) = S_{\text{original}}(x, y, z) \cdot (1 - M(x, y, z))\]

where $M(x, y, z)$ denotes the occluded section.

Fig.~\ref{fig:rai_example}(c) illustrates an instance of occlusion observed in LiDAR data.

\paragraph{iii. Weather Disturbances}
Adverse weather conditions such as heavy rain, fog, snow, or glare can introduce noise and distortions in sensor data, affecting the vehicle's ability to perceive its surroundings accurately. Raindrops on camera or LiDAR sensors can scatter light and cause reflections, leading to blurred or obscured images. CARLA simulator already provides a simulation of these different weather conditions, and we leveraged this in our experiment.

\subsection{Robustness against Sensor Errors}
Sensor noise resulting from hardware faults, calibration issues, or sensor drifts can lead to erroneous readings and misinterpretations of the environment, consequently impacting critical tasks such as obstacle detection, and planning. Detecting and mitigating sensor errors is crucial for ensuring the robustness and accuracy of AD systems.
\paragraph{i. Camera Error}
Errors associated with faulty camera sensors, dead sensor elements, inaccurate pixel interpretation process, and intermittent electrical interference result in poor camera outputs (modelled with salt and pepper noise). This type of noise manifests as sparsely occurring white and black pixels in images, affecting the visual quality and integrity of the captured data. High-amplitude intermittent electrical interference, such as arcing on electrical contacts, can also contribute to the occurrence of salt and pepper noise in camera outputs.
Formally, we add salt and pepper noise to each pixel independently. The resulting camera image \(I_{\text{noisy}}(x, y) \) is defined as:

$$ I_{\text{noisy}}(x, y) = 
\begin{cases} 
0 & p \\
255 & {1 - p} \\
I_{\text{original}}(x, y) & \text{otherwise}
\end{cases} $$
Where $p$ is the probability that an arbitrary pixel would have a value of 0.
Understanding the sources of salt and pepper noise is crucial for developing effective noise removal techniques to enhance the quality of images captured by cameras in AD systems.

\paragraph{ii. LiDAR Error}
3D LIDARs are incorporated with multiple channels to enable a higher vertical field of view (FOV), where each channel is a distinct laser beam. One example of hardware failure in 3D LIDARs is channel failure, where a particular beam stops functioning. To simulate such a fault, we remove sensor readings corresponding to arbitrarily selected channels. See examples of noisy and faulty sensor outputs in Fig.~\ref{fig:rai_example}.

\paragraph{iii. Other Sensors Errors}
In addition to cameras and LiDARs, we introduced random noise, drawn from a uniform distribution, to the sensor data coming from positional measurement-related sensors such as; the Global Navigation Satellite System (GNSS), the Inertial Measurement Unit (IMU), and the speedometer. These sensors are naturally susceptible to faults due to hardware limitations, signal interference, or environmental conditions. Hence, assessing driving agents' robustness along these dimensions is important.

Given sensor data $S$, noised sensor data $\hat{S}$, is obtained as:
\[\hat{S} = S + n ~;~n \sim U(-N, N)\]
where $n$ represents added noise, $U$ denotes uniform distribution, and $N$ controls the magnitude of the noise.

\subsection{Robustness against Corner Cases}
\paragraph{i. Corner Cases}
Corner cases represent extreme situations that may not be encountered in regular traffic but have the potential to challenge the capabilities of AD systems. This includes situations where an actor violates traffic rules, e.g., a pedestrian crossing outside of a crosswalk, debris on the road, etc.

It could also be the data drift effect which occurs when there's a divergence between the test (or production) data distribution and the train data distributions. This potentially leads to model degradation and decreased accuracy. This could be a collective change in the behaviour of actors within a region, e.g., from being conservative to aggressive as the population density of actors increases in the region. It could also be the fading away of traffic signs. By monitoring data drift and implementing strategies to adapt models to changing data patterns, AD systems can maintain their effectiveness in diverse and evolving environments. Robustness can be ensured by utilising techniques such as drift detection, model re-calibration, and ongoing performance evaluation to ensure that the models remain robust and effective in the face of fluctuating environmental variables. We leverage CARLA's flexibility in defining environments and pedestrian behaviour to create a combination of these situations. We can create data shift scenarios if the details, e.g., the source of training data and its distribution are included in the datasheet.

\section{S-RAF: Carbon Emission Indicator}
\label{sec:codecarbon}
It is challenging to obtain an accurate measure of the overall $\text{CO}_2$ autonomous vehicles emits to the environment. However, we can estimate how much of $\text{CO}_2$ the AI model that powers the vehicles constitutes at inference time. We estimated this by obtaining the carbon intensity ($\text{CI}$) value for the region from which the model is run, and multiplying this by the amount of energy ($E$) the process running the model used up: \(\text{CO}_2\) emissions (in \(\text{Kg CO}_2\text{Eq.}) = \text{CI} \times {E}\).

The Carbon Intensity ($\text{CI}$) of the consumed electricity is calculated as a weighted average of the emissions from the different energy sources that are used to generate electricity, including fossil fuels and renewables. In the work by~\cite{schmidt2021codecarbon}, fossil fuels coal, petroleum, and natural gas are associated with specific carbon intensities based on a known standard amount of $\text{CO}_2$ emitted for each kilowatt-hour of electricity generated. This is based on publicly available charts. Renewable or low-carbon fuels include solar power, hydroelectricity, biomass, geothermal, etc. The nearby energy grid contains a mixture of fossil fuels and low-carbon energy sources, called the Energy Mix. Based on the mix of energy sources in the local grid, the Carbon Intensity of the electricity consumed is calculated. The nearby grid is determined based on the location of the compute resource. 

$E$ is the energy consumed by the computational infrastructure (both GPU and RAM) expressed in kilowatt-hours. See~\cite{schmidt2021codecarbon} for further implementation details.

\section{Experiment}
\label{sec:experiment}
\subsection{Traffic Setup}
The traffic setup was done in CARLA simulator~\cite{dosovitskiy2017carla}. For the experiment reported in this paper, we created a complicated route and spawned multiple actors to roam around the town. The route consists of different road structures including junctions and intersections (based on the NHTSA~\cite{NHTSA} topology). Actors include different forms of vehicles spanning cyclists to trucks interacting in different forms with the agent being tested. This traffic setup is similar to those used in the previous CARLA AD Challenge~\cite{leaderboard}.

\subsection{Driving Agent Details}
We selected three trained agents, LBC~\cite{chen2020learning}, NEAT~\cite{chitta2021neat}, and InterFuser~\cite{shao2023safety}, 
from the 2020, 2021, and 2022 CARLA Challenges, respectively based on the leaderboard results. We selected one agent from each year using two steps (i) sorting by driving scores (ii) selecting the top agent that provided enough details for easy reproducibility. 

The authors of the selected agents were quite transparent by providing model weights, code, and other materials useful for running S-RAF. We refer to this as \textit{process-based transparency}, where detailed information about the processes involved in the agent's development stage is made available. Other forms of transparency important for responsible autonomous driving include \textit{output-based transparency} where the agent provide human-understandable explanations for their predictions, and \textit{method-based transparency} which relates to the architecture used, e.g., modular or end-to-end~\cite{omeiza2021explanations}. A modular architecture is assumed to offer better transparency. We examined the available technical materials (e.g., code base, research paper, etc) of the selected agents to see how they align with these different transparency dimensions. As this is mainly a qualitative assessment and quite subjective, we do not consider it an indicator in S-RAF. See Table~\ref{tab:transparency} for details.

\subsection{Regular Driving Score}
We used the driving score metric to assess the driving performance of the agents. The driving score \(D_{i} = R_{i} P_{i}\) is the product of the route completion and the infraction penalty. Here \(R_{i}\) is the percentage of completion of the \(i-th\) route and \(P_{i}\) is the infraction penalty incurred during the \(i-th\) route.
We track different types of infractions (e.g., collision with a vehicle, running a red light, etc.) in which the agent was involved. The infraction penalty score aggregates all of the infractions as a geometric series along a particular route. Agents start with an ideal 1.0 base penalty score, which is reduced each time an infraction is committed. Infraction penalty $ P_{i}$ was computed as: \(\prod\limits_{l}^{|{W}|} {(P_{l})}^{\left|{W_l}\right|}\), where $W$ is a set of infractions, $l$ is an infraction type in $W$, and $|W_l|$ is the number of $W_l$ infractions that occurred, and $P_l$ is the cost for infraction type $l$.

\subsection{Robustness Driving Score}
We ran one route multiple times introducing each of the different types of disturbances in each run. For example, if the traffic disturbance was weather and there were three weather types, then the route would be run three times corresponding to the three different weather types. The run condition, in this case, is  `weather'. When a route has been run for the desired number of times:
\begin{enumerate}

    \item The driving scores (i.e., after penalties have been applied) for the runs are grouped based on the type of traffic disturbances introduced (condition). For example, weather driving scores [...], camera noise driving scores [...], ...

    \item The driving score for each type of traffic disturbance/condition $j$ is obtained by selecting the minimum score across all runs $k$ under condition $j$.
    from the runs under this condition: \[D^{j}_{i}={\underset{k}{\min} (D^{j}_{i, k})}\]. For example, if a route was run in $n$ different weather conditions, the driving score for condition weather would be the least driving score obtained after running the agent in the $n$ different weather conditions. In the situation that more than one routes were used, the score from the route that produced the lowest score for the selected traffic disturbance type would be used.
    
    \item  A robustness ratio is computed per condition. Robustness ratio for $j-th$ condition is the ratio of the driving score obtained when the disturbance is introduced and the driving score obtained in the normal/regular driving case where no disturbance was introduced, that is, \(s_j = \frac{D^{j}_{i}}{D_i}\)
    \item A robustness driving score for which agents are ranked is computed as by taking the mean of all $D^{j}_{i}$. \(RDS~=~\frac{1}{m} \sum_{j}^{m} D^{j}_{i}
\)
where $m$ is the total number of conditions.
\end{enumerate}

\subsection{Estimating Carbon Emissions}
For each run, we estimate the amount of carbon (in $\text{Kg CO}_2 \text{ Eq.}$) emitted as a result of running the agent. Note that only the system process running the agent is tracked. We provide the average $\text{CO}_2$ emission for the runs as well as the average $\text{CO}_2$ emission per second. The emissions estimation procedure is based on the codecarbon~\cite{schmidt2021codecarbon} implementation explained in the previous section.
\begin{figure*}[h]
    \centering
    \begin{minipage}[b]{0.48\textwidth}
        \centering
        \includegraphics[width=\textwidth]{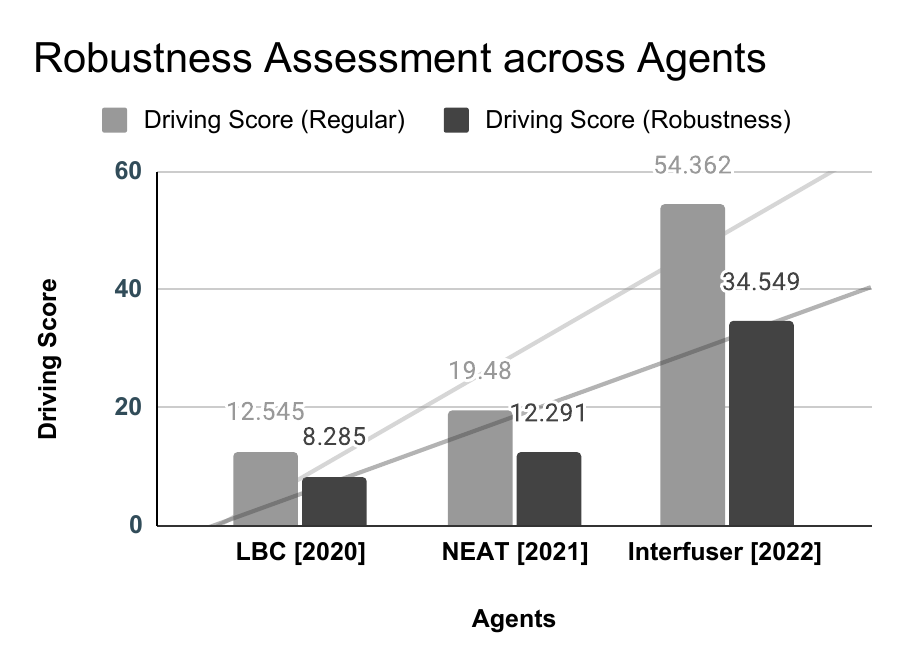}
        \caption{All agents' performance dropped with disturbances. CARLA's NPC Agent was excluded as it doesn't rely on sensor data.}
        \label{fig:robustness}
    \end{minipage}
    \hfill
    \begin{minipage}[b]{0.48\textwidth}
        \centering
        \includegraphics[width=\textwidth]{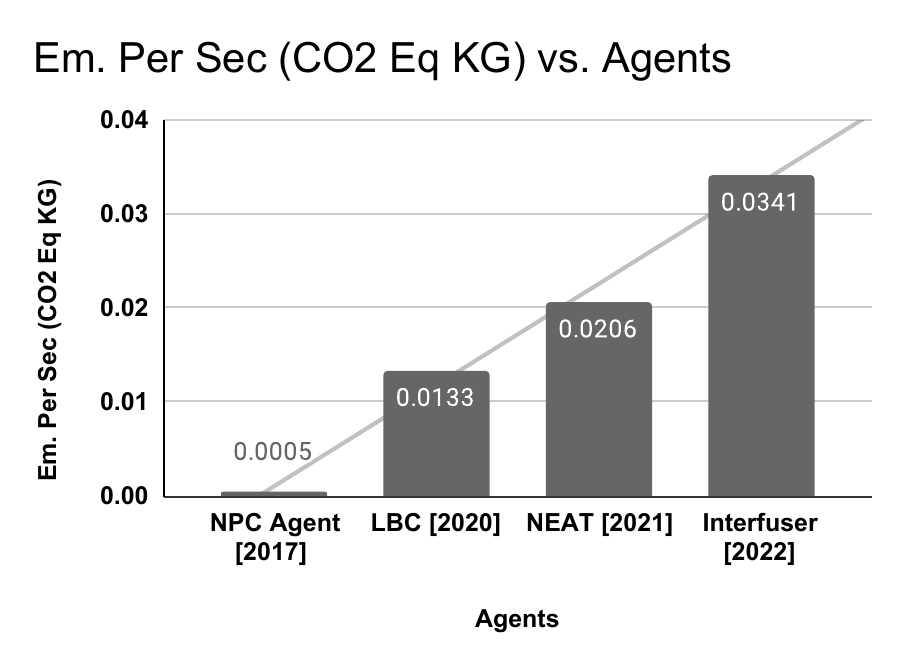}
        \caption{We observe increased emissions as the agents get more robust over the years, with NPC Agent constituting significantly low emissions.}
        \label{fig:emissions}
    \end{minipage}
\end{figure*}
\begin{table*}[!htb]
\centering
\caption{\small{Qualitative details on transparency of agents. Out.: Output; Mtd: Method; E2E: End-to-End; IC: Interpretability Considerations (i.e., whether authors implemented interpretability techniques); CD: Code Documentation; MW: Model Weights; MC: Model Card; DS: Datasheet; TR: Technical Report.}}\label{tab:transparency}
 \scriptsize
\begin{tabular}{lrrrrrrrrrrr}\toprule
& & &\multicolumn{7}{c}{\textbf{Process}} & \\\cmidrule{4-10}
\textbf{Model} &\textbf{Out. (2)} &\textbf{Mtd.} &IC &Code &CD &DS &MW &MC &TR \\\midrule
LBC &\ding{55} &E2E &\ding{55} &\ding{51} &\ding{51} &\ding{51} &\ding{51} &\ding{51} &\ding{51} \\
NEAT &\ding{55} &E2E &\ding{51} &\ding{51} &\ding{51} &\ding{51} &\ding{51} &\ding{51} &\ding{51} \\
InterFuser &\ding{55} &E2E &\ding{51} &\ding{51} &\ding{51} &\ding{51} &\ding{51} &\ding{51} &\ding{51} \\
\bottomrule
\end{tabular}
\end{table*}

\begin{table*}[!htp]\centering
\caption{\small{Performances of Previous CARLA Challenge Agents from 2020 to 2022, including CARLA'S NPC Agent. DS: Driving Score, RDS: Robustness Driving Score, AEPS: Average Emission Per Seconds ($\text{Kg CO}_2 \text{ Eq.}$), AEPR: Average Emissions Per Route ($\text{Kg CO}_2 \text{ Eq.}$), Cam.: Camera, CO: Camera Occlusion, LO: LiDAR Occlusion, Spdm: Speedometer, ARC: Average Route Completion (only on conditioned routes $D^{j}_{i}$), ASTPR: Average Simulation Time Per Route. Hyphens (-) means sensor is absent.}}\label{tab:rai-result}
\scriptsize
\begin{tabular}{lrrrrrr|rrrrr|rr|rrr}\toprule
&\multicolumn{6}{c}{\textbf{Robustness aganist Environmental Disturbances}} &\multicolumn{5}{c}{\textbf{Robustness against Sensor Faults}} &\multicolumn{2}{c}{\textbf{Env. Sustainability}} &\multicolumn{2}{c}{\textbf{Simulation Details}} \\\cmidrule{2-16}
\textbf{Models} &DS (\%) &RDS (\%) &CO &LO &Wth. &Drift &Cam. &LiD &GNSS &IMU &Spdm &AEPS &AEPR &ARC(\%) &ASTPR($s$) \\\midrule
NPC Ag. &26.697 &- &- &- &- &- &- &- &- &- &- &\textbf{0.0005} &\textbf{0.090} &38.139 &\textbf{189.300} \\
LBC &12.545 &8.285 &0.241 &- &0.216 &\textbf{0.999} &\textbf{0.472} &- &\textbf{0.999} &0.693 &\textbf{0.999} &0.0133 &4.636 &28.222 &336.085 \\
NEAT &19.48 &12.291 &0.145 &- &\textbf{0.723} &0.978 &0.001 &- &0.57 &\textbf{0.999} &0.999 &0.0206 &7.468 &32.932 &356.936 \\
Interfuser &\textbf{54.362 }&\textbf{34.549} &\textbf{0.932} &0.29 &0.239 &\textbf{0.999} &0.239 &0.804 &0.546 &\textbf{0.999} &0.668 &0.0341 &12.418 &\textbf{52.188} &367.017 \\
\bottomrule
\end{tabular}
\end{table*}

\label{sec:results}

\section{Results}
\subsection{Robustness}
From Table~\ref{tab:rai-result}, we observe that Interfuser had the highest driving score and overall robustness driving score. This is not surprising as it has the highest number of sensors for accurate perception and planning. Camera seems to be the most intolerant to faults as we observe the greatest decline when the camera was noised. Cameras are very important sensors for navigation. In fact, some AD companies are beginning to only rely on cameras for navigation. Robustness readings need to be put in perspective with average route completion (ARC) and driving score. For instance, while LBC seems to have impressive robustness scores for many of the runs, it should be noted that the agent was only able to complete 28\% of each route on average. Its driving score is as well low. This information tells us that irrespective of the impressive robustness scores for many of the sensors, LBC doesn't seem to be a safe and performant agent.  

\subsection{Carbon Emissions}
It is challenging to assess how much emission a vehicle in its entirety contributes to the environment. However, we can estimate how much emission the underlying AI model that controls navigation constitutes. Hence, we estimated the amount of $\text{CO}_2$ emitted when the AI model is run over time. For fair comparisons, we estimated average emissions per second (AEPS). From Fig.~\ref{fig:emissions} and Table~\ref{tab:rai-result}, we see that the default CARLA NPC Agent that uses the classical navigation algorithms (non-machine learning) constitutes the lowest $\text{CO}_2$ emission. This agent doesn't utilise sensors, it rather uses ground truth information for planning. Hence, no robustness scores were assigned. The agent was useful to benchmark $CO_2$ emissions. As AD agents get more sophisticated (with increased robustness), the amount of $\text{CO}_2$ they emit increases.

\section{Discussion and Limitations}
\label{sec:discussion}
We have drawn attention to the need for accessible frameworks with RAI indicators, starting with robustness. These frameworks serve as safety guardrails for AD agent development. We took a socio-technical approach in this work by first establishing the importance and the need for robustness indicators and then proposed a technical framework that offers metrics for robustness and environmental sustainability assessment (using $\text{CO}_2$ emission as proxy). Our framework (S-RAF) focused on streamlining AD agents development process rather than the purpose for which the agent is developed. While the Responsible Research and Innovation (RRI) frameworks (e.g., AREA framework~\cite{jirotka2017responsible} underscore the importance of the purpose inputted into a machine, assessing such purposes is nearly intractable. S-RAF thus focuses on enhancing development processes to achieve safe and responsible driving agents. S-RAF improves over the conventional AI agents assessment approach that focuses on improving prediction accuracy with limited examples of critical edge cases by quantifying different aspects of robustness and environmental sustainability.

S-RAF's robustness metric takes into account core safety critical sensors in an AV and questions the capability of the agent when an individual or a combination of these sensors malfunctions or when their sensing range is limited due to environmental factors. These cases are hard to obtain in the real world. But with S-RAF, we can simulate these cases affordably. Our results indicate an increasing trend in robustness and safe driving capability over the years. With the breakthrough in generative AI, where generalisable models are being developed, we believe that AD agents would witness a tremendous increase in capabilities. However, we advocate RAI principles and the use of S-RAF in the process.

The emission of green gases by vehicles have impacts on human safety~\cite{ogur2014effect}. With S-RAF, AI developers have the option to discard models that emit excessive amounts of $\text{CO}_2$. The limitation with S-RAF regarding sustainability is that it does not factor in the emissions caused during the manufacturing process of hardware components, e.g., electric vehicle batteries~\cite{ji2012electric} which are notable for constituting $\text{CO}_2$ emissions at manufacturing time.
We saw that $\text{CO}_2$ emissions increase as the agents get more performant. No surprises as this adheres to the scaling law (the larger the better). Should the development of extremely large models be stopped? This is an ongoing debate. But again, we encourage developers to consider RAI principles and also optimise for low S-RAF $\text{CO}_2$ emissions.

We have only tested S-RAF on synthetic driving data from CARLA, this is one limitation of this work. Also, when S-RAF assesses agents, it indirectly emits $\text{CO}_2$ as it runs the agents. However, this run is only for a limited time compared to when the models are already deployed and run for an extended period. Thus, S-RAF's use is justifiable as it helps to prevent the deployment of unsafe models. Another limitation of this work is that we have only tested on a handful of AD agents due to the scarcity of open-source AD agents. Hence, we have created a challenge website\footnote{https://carla-rai-challenge.github.io/} that encourages developers to submit their agents for assessment and receive feedback for improvements. This challenge would yield more examples from which we can further validate S-RAF while respecting the privacy and copyright agreements of the agent/model owners. That said, we encourage developers and corporate entities to embrace all transparency dimensions highlighted in this paper (process-based, method-based, and output-based) to facilitate easy assessment and potentially, increased safety.

\section{Conclusion}
\label{sec:conclusion}
We have argued for the need for an accessible framework for assessing AD agents. We developed S-RAF, a framework aimed at guiding developers in building responsible AD agents that are robust and environmentally friendly. S-RAF can equally support AV regulators and other authorised stakeholders in assessing AD agents. Through an experiment, we showed how indicators composed in S-RAF were developed, and we tested these indicators on benchmark AD agents in CARLA simulator. We saw improvements in robustness over time (from 2020 to 2022), while the amount of $\text{CO}_2$ emissions has increased as the models got sophisticated over the years. Lastly, we discussed the implications of these findings and future research agenda.
\section{Acknowledgement}
This work was supported by the EPSRC RAILS project
(grant reference: EP/W011344/1), Amazon Web Services (AWS), and the Embodied AI Foundation.
\small{
\bibliography{aaai24} 
}
\end{document}